\documentclass{article}
\usepackage{spconf,amsmath,epsfig}
\usepackage{algorithm}
\usepackage{algpseudocode}
\usepackage{amsmath}
\usepackage{graphics}
\usepackage{epsfig}

\usepackage{graphicx}
\usepackage{amsmath}
\usepackage{amssymb}
\usepackage{booktabs}
\usepackage{times}
\usepackage{epsfig}

\usepackage[capitalize]{cleveref}
\crefname{section}{Sec.}{Secs.}
\Crefname{section}{Section}{Sections}
\Crefname{table}{Table}{Tables}
\crefname{table}{Tab.}{Tabs.}

\let\OLDthebibliography\thebibliography
\renewcommand\thebibliography[1]{
  \OLDthebibliography{#1}
  \setlength{\parskip}{0pt}
  \setlength{\itemsep}{0pt plus 0.3ex}
}

\begin{document}\sloppy

\def\x{{\mathbf x}}
\def\L{{\cal L}}

\title{MISS: Memory-efficient Instance Segmentation Framework \\By Visual Inductive Priors Flow Propagation}
%

\name{Chih-Chung Hsu*, Chia-Ming Lee}
\address{Institute of Data Science, National Cheng Kung University, Taiwan\\
cchsu@gs.ncku.edu.tw*, zuw408421476@gmail.com}

\maketitle

\begin{abstract}
Instance segmentation, a cornerstone task in computer vision, has wide-ranging applications in diverse industries. The advent of deep learning and artificial intelligence has underscored the criticality of training effective models, particularly in data-scarce scenarios—a concern that resonates in both academic and industrial circles. A significant impediment in this domain is the resource-intensive nature of procuring high-quality, annotated data for instance segmentation, a hurdle that amplifies the challenge of developing robust models under resource constraints.
In this context, the strategic integration of a visual prior into the training dataset emerges as a potential solution to enhance congruity with the testing data distribution, consequently reducing the dependency on computational resources and the need for highly complex models. However, effectively embedding a visual prior into the learning process remains a complex endeavor.
Addressing this challenge, we introduce the MISS (Memory-efficient Instance Segmentation System) framework. MISS leverages visual inductive prior flow propagation, integrating intrinsic prior knowledge from the Synergy-basketball dataset at various stages: data preprocessing, augmentation, training, and inference. Our empirical evaluations underscore the efficacy of MISS, demonstrating commendable performance in scenarios characterized by limited data availability and memory constraints.
\end{abstract}

\section{Introduction}
\label{sec:intro}

\begin{figure}
    \centering
    \includegraphics[width=0.48\textwidth]{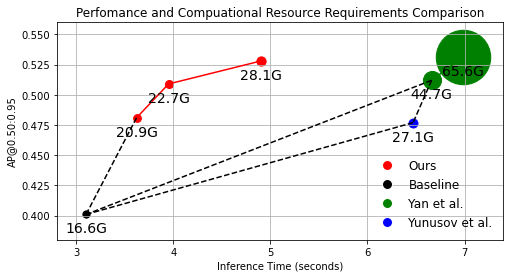}
    \caption{Comparison of performance and computational resource requirements between our approach and previous benchmark \cite{2021sota}\cite{2022sota}, in Synergy-basketball dataset. Compared with other methods, our method significantly reduces the demand for computational resources. In the figure, the size of the circles represents the memory usage for each method.}
    \label{fig:exp1.png}	
\end{figure}

With the advent of deep learning, industries are increasingly integrating it into various facets of their operations to secure a competitive advantage. However, the implementation of deep learning at detailed levels encounters challenges, including limited annotated data and computational resources. These constraints often result in suboptimal model performance. Consequently, the effective use of scarce data for model training has emerged as a critical research area. The traditional approach involves expending additional human resources to gather and annotate more extensive data sets and designing complex model architectures, thereby imposing further burdens.

In contrast, this work introduces "Visual Inductive Priors Flow Propagation" - a robust, prior-guided, data-aware augmentation pipeline. This method effectively integrates abundant priors into various understanding tasks that require high-quality, large-scale data for fine-grained image or video understanding, such as instance segmentation and multi-object tracking. By fully leveraging prior knowledge from datasets or background rules, our approach reduces the need for extensive data and computational resources.

To validate our proposed method, we employ a basketball-type sports instance segmentation as our benchmark. Image understanding in sports scenarios must account for factors like varying shooting angles, camera dynamics, rapid object movement, diverse lighting conditions, and scene complexity. These elements make achieving high-precision prediction results challenging, especially with severely limited training data or computational resources.

However, sports scenarios offer a wealth of prior knowledge, rooted in well-established background rules. This includes valuable information such as the layout of a basketball court, regulations for player attire, and distinctive team jersey designs. We posit that integrating these inductive priors into the training process allows the model to achieve highly granular segmentation results during inference. Employing this strong prior knowledge, incorporated into the dataset during the offline phase, enhances model convergence during training without necessitating significant computational resources.


In this paper, we introduce the Synergy-basketball dataset, courtesy of Synergy Sports. This dataset annotates individuals and basketballs within a basketball court context, encompassing categories such as referees, players, and coaches. Our experimental results demonstrate the efficacy of our model, which can be trained on a single GPU with 24 GB of memory, while still delivering notable performance. The principal innovations and contributions of this work are threefold:

\begin{itemize}
\item To the best of our knowledge, this is among the first works to propose a streamlined, memory-efficient learning framework. This framework effectively incorporates a visual prior for instance segmentation tasks, proving particularly beneficial in scenarios with limited data and computational resources.
\item We explore the concept of a visual prior, inherent in the dataset or derived from background rules, and apply this knowledge comprehensively across both online and offline phases. This includes data preprocessing, training, and inference.
\item Our proposed method performs better than existing deep instance segmentation models, specifically in the Synergy-basketball dataset. It achieves this with reduced memory usage and runtime complexity, without sacrificing performance.
\end{itemize}

\begin{figure*}
    \centering
    \includegraphics[width=0.98\textwidth]{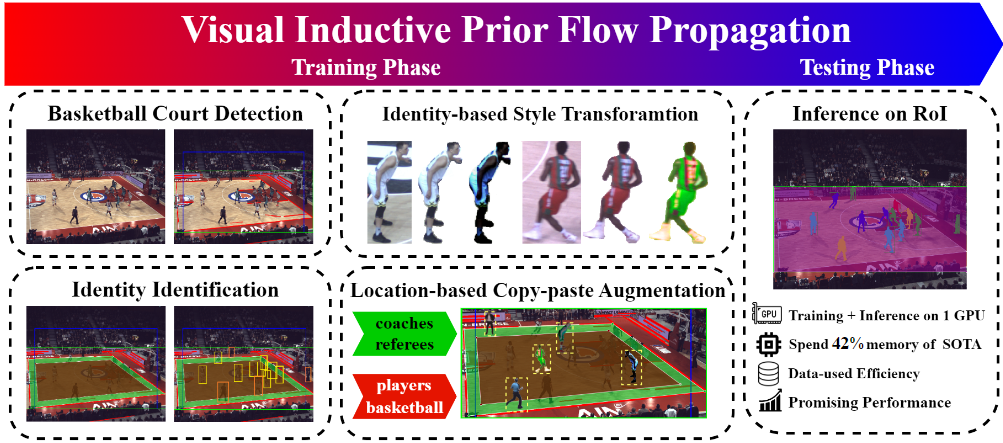}
    \caption{The overall  of proposed instance segmentation framework. The visual inductive prior is fully utilized at each stage to make effective optimizations. This approach not only reduces computational resource consumption but also maintains solid model performance. We begin by employing the Canny-Hough operator to adaptively combine image-level prior to detect the basketball court's position. Subsequently, we leverage class-level prior for identity identification. We then utilize this information for style transformation of various objects, integrating image-level prior knowledge through copy-paste augmentation. Finally, model inference solely based on the detected basketball court's location.}
    \label{fig:overall.png}	
\end{figure*}

\section{Related Works}

In Section 2, we present a comprehensive overview of prevalent model architectures utilized in recent years for instance segmentation tasks, complemented by a discussion on data augmentation techniques that bolster prediction performance.

Instance segmentation methods based on deep learning have achieved significant recognition. This is illustrated by methods such as Cascade-MaskRCNN \cite{CascadeRCNN}, Maskformer \cite{Maskformer}, QueryInst \cite{QueryInst}, and CBNet \cite{CBNet}. Recent studies \cite{2021sota}\cite{2022sota} have showcased outstanding performances on the Synergy-basketball dataset using CBNet-based model architectures.

CBNet stands out for its ability to effectively combine multiple backbone networks and detectors. This synergy facilitates the integration of both low-level and high-level semantic information. The architecture is notable for its scalability and ease of training, thereby conferring a broad spectrum of inductive capabilities to the model. It achieves this while preserving significant detection accuracy and generalization capabilities, without detriment to inference speed.

Data augmentation is a pivotal aspect in deep learning, particularly in scenarios of limited data availability. This improvement bolsters the model's ability to generalize across novel domains. An effective augmentation strategy for instance segmentation tasks is Copy-paste augmentation \cite{CopyPaste}, which utilizes prior object knowledge to strengthen model generalization and robustness against out-of-domain objects.

In previous studies on the Synergy-basketball benchmark \cite{VIP2th}\cite{VIP3th}, Yunusov et al. introduced task-specific augmentation \cite{2021sota}, constraining object placement in copy-paste augmentation to a predefined area. Yan et al. \cite{2022sota} proposed integrating additional prior knowledge, enabling the object placement range to adaptively adjust in response to the camera pose.  However, the latter method necessitated extensive post-processing, including significant image enlargement and testing-time augmentation, consequently increasing computational resource requirements.


\section{Methodolodgy}

In this section, we elaborate on all the components presented in Figure \ref{fig:overall.png}, including the basketball court detection algorithm, illustration of augmentation pipeline, and inference on the region of interests.

\label{sec:methodology}

\subsection{Basketball Court Detection and Cropping}

Handling images of substantial size often results in extended training and inference durations and can lead to memory constraints. A common strategy to mitigate this issue is resizing images to a uniform dimension. However, this approach might distort salient features or lose crucial texture details in the images. An alternative method, cropping, tends to preserve more information from the original images. This technique relies heavily on prior knowledge embedded within the image to ascertain precise cropping boundaries. In this work, we propose a novel algorithm for basketball court detection and cropping. This algorithm employs a Canny-Hough straight line detection operator \cite{Canny}\cite{Hough} to accurately identify the location of the basketball court, thereby facilitating the reduction of image size without significant loss of relevant information.

\begin{algorithm}
\caption{Basketball Court Detection Algorithm}
\begin{algorithmic}[1]
\State \textbf{Data:} All image data denoted as $\textbf{I}_{\text{original}}$, All Cropped image data denoted as $\textbf{I}_{\text{cropped}}$
\State Denote $\phi(\cdot)$ as canny operator, and $\tau(\cdot)$ as hough operator
\For{each image $\textbf{I}_i$ in $\textbf{I}_{\text{original}}$}
    \State Initialize $\textbf{I}_{\text{ih}}$, $\textbf{I}_{\text{iw}}$ = the height and width of image $\textbf{I}_i$
    \State $\text{min}_{\text{h}} = \frac{1}{9} \textbf{I}_{\text{ih}}$, $\text{max}_{\text{h}} = \frac{8}{9} \textbf{I}_{\text{ih}}$, \State$\text{min}_{\text{w}} = \frac{1}{15} \textbf{I}_{\text{iw}}$, $\text{max}_{\text{w}} = \frac{14}{15} \textbf{I}_{\text{iw}}$
    \State Detect all lines $L$ in images $\tau(\phi(\textbf{I}_i))$
    \State Compute the maximum convex hull $\delta$ contained in $L$
    \State Crop $\textbf{I}_i$ based on the coordinate $(x, y, w, h) =$\\
    \hspace{\algorithmicindent}(min($\text{min}_{\text{w}}$, $\delta_x$), max($\text{min}_{\text{h}}$, $\delta_y$) - 50,\\
    \hspace{\algorithmicindent}max($\text{min}_{\text{w}}$, $\delta_w$), min($\text{max}_{\text{h}}$, $\delta_h$))
\EndFor
\end{algorithmic}
\end{algorithm}

\begin{figure}
    \centering
    \includegraphics[width=0.48\textwidth]{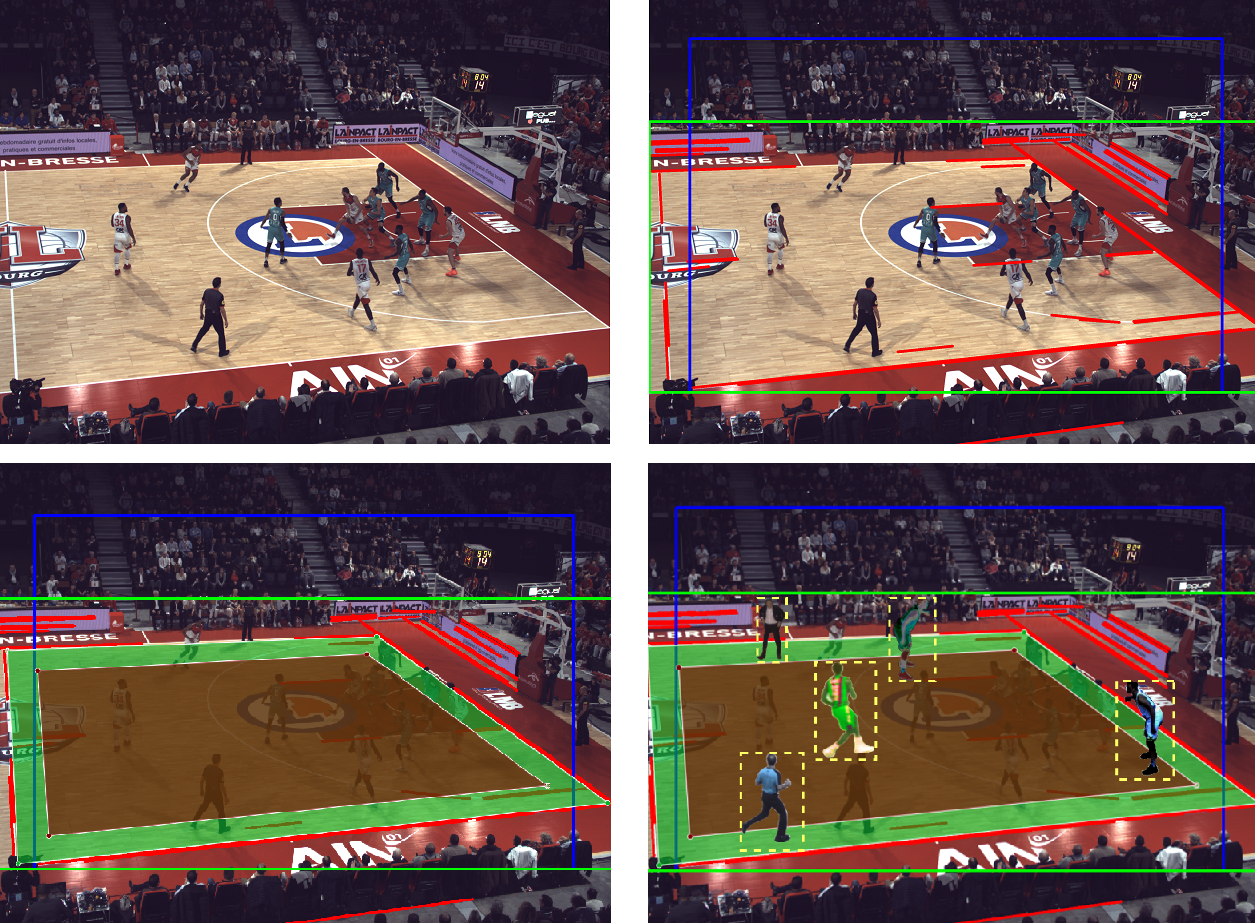}
    \caption{The illustrations for the cropping algorithm and location-based copy-paste augmentation. The top-left figure is the original image. The top-right one is cropped, with red lines detected by the Canny-Hough operator. The blue line shows a boundary based on image size, while the green lines indicate dynamic boundary from the detected lines. The two picture below display a region identified based on the maximum convex hull, which is determined using the endpoints of all lines detected by the Canny-Hough operator. The object marked by a dotted line is pasted . The subclass attributes of the object are determined by its bounding box coordinates.}
    \label{fig:basketball detection and cropping.png}	
\end{figure}

\begin{figure*}
    \centering
    \includegraphics[width=0.98\textwidth]{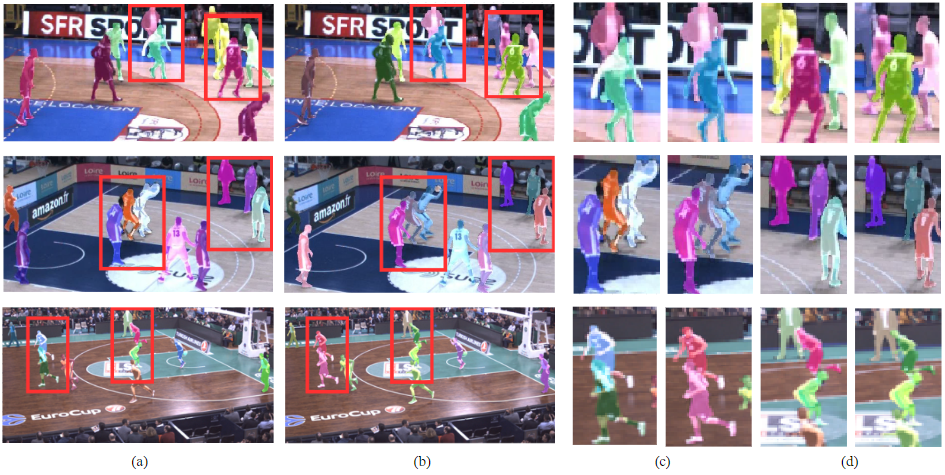}
    \caption{The results visualization of proposed method. (a) is the result of using simple copy-paste augmentation. (b) showcases the results using our proposed method. During the experiments, to ensure the simplicity of the methods for easy comparison, we excluded any post-processing or testing-time augmentation mentioned in this paper. (c) and (d) provide a magnified comparison of the areas in the images where artifacts are generated.}
    \label{fig:visual.png}	
\end{figure*}

\subsection{Enhancing Data Augmentation through Prior Knowledge Utilization}

In our endeavor to refine the data augmentation process, leveraging prior knowledge emerges as a pivotal strategy. Our analysis reveals a distinct spatial distribution in basketball games: referees and coaches predominantly occupy the court's perimeter for an unobstructed view, whereas players are primarily active within the court's core.

Exploiting this spatial pattern, we propose an enhanced method for object identity estimation based on the detection of areas within the basketball court. We designate 20\% of the detected area as a critical decision boundary for object identification. This approach enables a more targeted and accurate determination of object identities, essential for optimizing our data augmentation pipeline.


\subsection{Identity-Based Style Transformation}

In the realm of data augmentation, existing methodologies often confront two predominant challenges: (1) Augmentations at the whole-image level might inadvertently escalate the complexity of the feature space, thereby yielding marginal performance enhancements. (2) The augmentation techniques tailored for distinct classes or scenarios frequently appear inadequate.

Considering objects on a basketball court as a case in point, the diversity is apparent. The 'people' category, for instance, encompasses various sub-classes such as 'player', 'referee', and 'coach', each characterized by a rich array of internal feature attributes. Similarly, the 'ball' category exhibits variations under different lighting conditions and degrees of occlusion. The conventional approach of using a limited number of classes to define object attributes falls short in effectively augmenting the source domain data, as it oversimplifies the complexity inherent in these categories. An incremental decomposition of these high-level classes, informed by prior knowledge, can significantly enhance the model's aptitude in recognizing and adapting to unseen targets.

More specifically, our approach involves identifying the sub-classes of objects as delineated in Section 3-2 and subsequently applying distinct enhancement strategies to each. For the 'player' sub-class, we engage in object-level data augmentation through RGB curve distortion. This technique is particularly effective in accentuating the distinctiveness of players, who may don attire in varying colors, exhibit high saturation levels, and display stark contrasts. Additionally, variations in skin tone and gender among players can also be taken into account.

In contrast, for categories such as referees, coaches, and balls, where the scope of prior knowledge is relatively constrained, we employ salt-and-pepper noise and brightness adjustments. These methods aim to bolster the model's resilience in the face of diverse lighting conditions.
\begin{figure}
    \centering
    \includegraphics[width=0.48\textwidth]{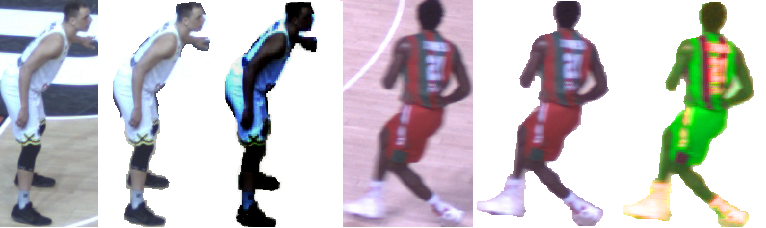}
    \caption{The demo of identity-based style transfer applied to basketball players. Significant variations in appearance are evident after the hue or RGB transformation. In the left example, there is a noticeable change in skin tone, while in the right example, the player's jersey changes dramatically, almost as if he has switched to a different team.}
    \label{fig:identity transform.png}
\end{figure}

\subsection{Location-Based Copy-Paste Augmentation}

Object coordinates in prior work involving copy-paste augmentation are limited to a range defined by the image's dimensions, indicated by a blue bounding box in Fig. \ref{fig:basketball detection and cropping.png}. This constraint can lead to objects being augmented in implausible locations. Our approach modifies these boundaries using prior knowledge. As detailed in Sections 3-2 and 3-3, our algorithm effectively identifies likely areas for different object types. Copy-paste augmentation is then executed within these tailored locations.

\subsection{Inference on Regions of Interest}

To optimize memory utilization and computational efficiency, our method focuses on inferring specific regions, as outlined in Sections 3-2 and 3-4. This approach involves cropping out extraneous areas using a vision-based prior. By resizing images to maintain only the regions of interest, we significantly reduce memory usage and inference time during model processing.

\section{Experiments}
\label{sec:experiment}

\subsection{Training Details}
Our experiments were conducted on a single NVIDIA TITAN RTX GPU, utilizing the MMDetection toolbox \cite{mmdetection}. We used the "Synergy-basketball" dataset, provided by Synergy Sports, comprising 184, 62, and 64 images in the training, validation, and testing sets, respectively.

In alignment with \cite{2022sota}, we maintained a consistent model architecture to ensure fair comparison. However, we opted for different hyperparameters in training, post-processing, and testing-time augmentation to enhance performance and memory efficiency. Our model underwent 36 epochs of training, employing the AdamW optimizer with a learning rate of 0.0001 and a weight decay of 0.05. Given computational resource constraints, the batch size was set to 1. Prior to training, training and validation images were duplicated tenfold and processed through our proposed data-preprocessing framework, integrating visual inductive priors.

\begin{figure}
    \centering
    \includegraphics[width=0.5\textwidth]{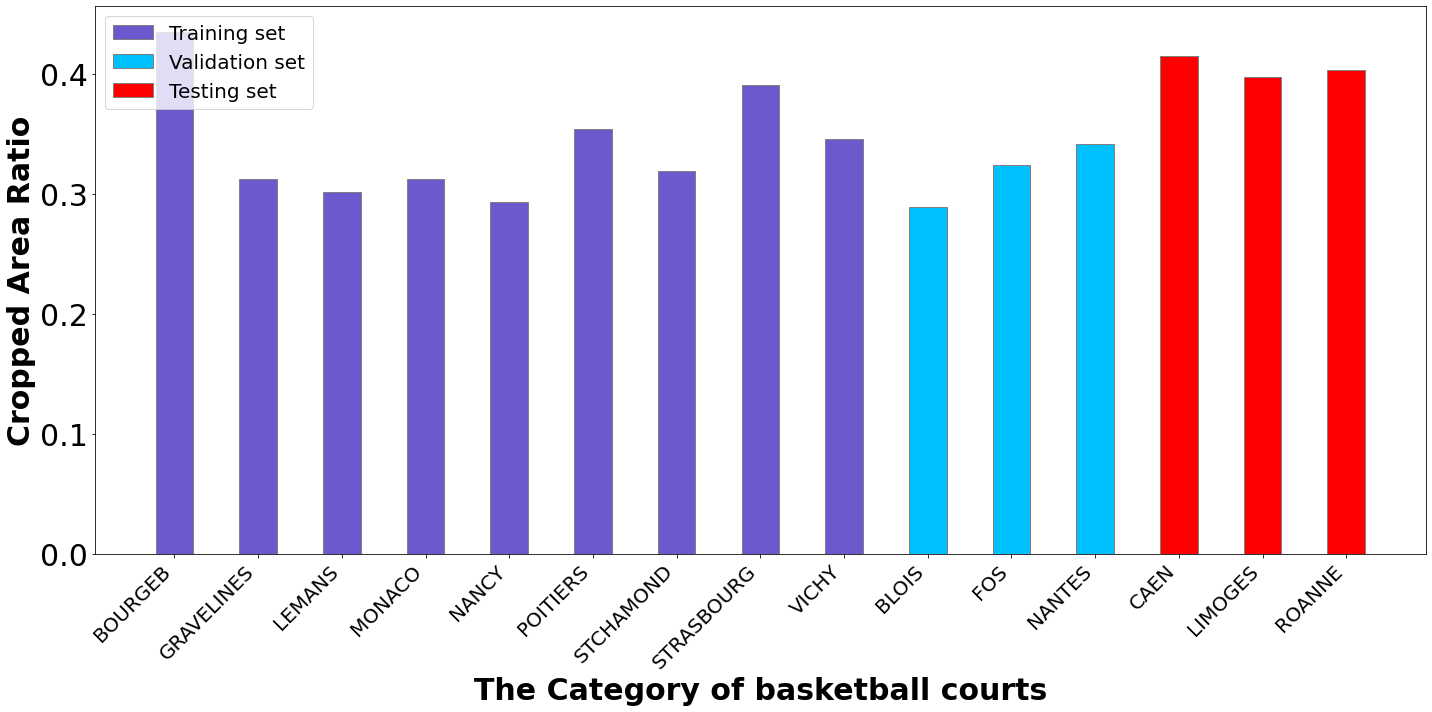}
    \caption{The cropped area statistic bar chart. The x-axis corresponds to basketball courts; the y-axis is the cropped area ratio against the whole raw image. From left to right, the three colors correspond to the training, validation, and testing set.}
    \label{fig:cropaoutcome.png}	
\end{figure}

Post-cropping, the image sizes in the training, validation, and testing sets were reduced to 66.02\%, 66.83\%, and 59.28\% of their original sizes, respectively. We also performed statistical analyses on each basketball court category, as depicted in Figure \ref{fig:cropaoutcome.png}.


\begin{table*}[]
\label{tab:ablationstudy}
\begin{tabular}{@{}lclclclclcl@{c}}
\toprule
\multicolumn{1}{c}{Models} & \multicolumn{1}{c}{AP@0.50}                        & \multicolumn{1}{c}{AP@0.50:0.95} & \multicolumn{1}{c}{\begin{tabular}[c]{@{}c@{}}AP@0.50:0.95\\ (small)\end{tabular}} & \multicolumn{1}{c}{\begin{tabular}[c]{@{}c@{}}AP@0.50:0.95\\ (medium)\end{tabular}} & \multicolumn{1}{c}{\begin{tabular}[c]{@{}c@{}}AP@0.50:0.95\\ (large)\end{tabular}} \\ \midrule
Vanillia instance segmentation model
&\multicolumn{1}{c}{0.789}&\multicolumn{1}{c}{0.403}&0.401&\multicolumn{1}{c}{0.470}&\multicolumn{1}{c}{0.631}\\ 
With simple copy-paste augmentation
&\multicolumn{1}{c}{0.863}&\multicolumn{1}{c}{0.444}&\multicolumn{1}{c}{0.462}&\multicolumn{1}{c}{0.561}&\multicolumn{1}{c}{0.667}\\ 
With proposed augmentation pipeline
&\multicolumn{1}{c}{0.870}&\multicolumn{1}{c}{0.481}&\multicolumn{1}{c}{0.515}&\multicolumn{1}{c}{0.579}&\multicolumn{1}{c}{0.700}\\ 
With post-processing &\multicolumn{1}{c}{0.896}&\multicolumn{1}{c}{0.509}&\multicolumn{1}{c}{0.533}&\multicolumn{1}{c}{0.584}&\multicolumn{1}{c}{0.731}\\ 
With testing-time augmentation&\multicolumn{1}{c}{0.924}&\multicolumn{1}{c}{0.528}&\multicolumn{1}{c}{0.558}&\multicolumn{1}{c}{0.601}&\multicolumn{1}{c}{0.742}\\ \bottomrule
\end{tabular}

\caption{Results of Ablation Study for the Synergy-basketball dataset with or without the proposed augmentation pipeline.}
\end{table*}

\subsection{Post-Processing and Testing-Time Augmentation}

Our proposed method, as detailed in Table \ref{tab:ablationstudy}, leverages prior knowledge and requires minimal resources and training data, achieving notable coarse-level segmentation results even without pre-trained models. To further enhance performance for specific application scenarios, we have developed additional post-processing and testing-time augmentation techniques. These enhancements, while improving model performance, also increase the computational resource requirements.

We adopt the Stochastic Weight Averaging (SWA) strategy \cite{zhang2020swa} to compute average model weights across training epochs. Additionally, we integrate variable-intensity GridMask augmentation and other techniques to enrich the training dataset with diverse samples. To further boost performance, we employ model ensemble strategies and the model-soup technique \cite{modelsoup}. Finally, to improve results, we increase the testing image size, cropping it based on basketball-court boundaries, to larger dimensions, such as 3400x2400 pixels.

\subsection{Ablation Study and Performance Comparison}

The results of our ablation study and performance comparisons are summarized in Tables \ref{tab:ablationstudy} and \ref{tab:comparison}. Our method, which incorporates a vision inductive prior, significantly outperforms traditional approaches.

In the AP@0.50 metric, our model demonstrates robust performance, effectively detecting most instances in the test set. While the AP@0.50:0.95 metric shows no substantial differences compared to the state-of-the-art method \cite{2022sota}, our model is notably more memory-efficient. When employing test-time augmentation, it uses only 42.1\% of the memory required by \cite{2022sota}, while maintaining competitive performance.

\begin{table}[]
\resizebox{\columnwidth}{!}{%
\begin{tabular}{lcccc}
\hline
\multicolumn{1}{c}{Methods} &
  \begin{tabular}[c]{@{}c@{}}AP@\\ 0.50:0.95\end{tabular} &
  \begin{tabular}[c]{@{}c@{}}AP@\\ 0.50\end{tabular} &
  \begin{tabular}[c]{@{}c@{}}Memory \\ (G)\end{tabular} &
  \begin{tabular}[c]{@{}c@{}}Inference times\\ (s)\end{tabular}\\ \hline
Yunusov et al.\cite{2021sota} & 0.477          & 0.747          & 27.1           & 6.47 \\
Yan et al. \cite{2022sota}    & 0.512 & 0.816     & 44.7            & 6.66\\

Yan et al. with TTA  & \textbf{0.531}         & 0.837 & 65.6 & 6.98\\

Our    & 0.509 &\textbf{ 0.896}     & \textbf{22.7}            & \textbf{3.95}\\\hline

\textbf{Ours with TTA}  & \textbf{0.528}          & \textbf{0.924} & 28.1  & 4.91

\end{tabular}%
}
\caption{Comparison of performance and computational resource requirements. *The architectures of these methods are CBNet-based, but the detailed training process or hyperparameters may differ.}
\label{tab:comparison}
\end{table}

\section{Conclusion}
\label{sec:conclusion}
This work introduces "MISS," an innovative framework for instance, segmentation, which innovatively incorporates visual inductive priors at multiple stages, encompassing data preprocessing, copy-paste augmentation, and model training and inference phases. Our experimental results reveal that this methodology significantly bolsters model performance in environments constrained by resources, without reliance on pre-trained weights or transfer learning techniques, even when minimal training data is available.
The escalating trend towards developing larger models in deep learning has precipitated an increased demand for computational resources. Conversely, there is a growing emphasis on training potent models and extracting representative features with sparse data. We posit that the utility of our research transcends the confines of sports analytics. The versatility of our "MISS" framework makes it a potent tool in any field replete with prior data, and its potential extends to other areas of computer vision, including object recognition and multi-object tracking.

\bibliographystyle{IEEEbib}
\bibliography{icme2022template}

\end{document}